\documentclass[conference]{IEEEtran}
\IEEEoverridecommandlockouts
\usepackage{url}

\usepackage{cite}
\usepackage{amsmath,amssymb,amsfonts}
\DeclareMathOperator*{\argmin}{arg\,min} 
\usepackage{algorithm}
\usepackage{algpseudocode}
\usepackage{comment}
\usepackage[caption=false, font=footnotesize]{subfig}
\usepackage{bbm}
\usepackage{graphicx}
\usepackage{svg}
\usepackage[normalem]{ulem}
\usepackage{textcomp}
\usepackage{xcolor}
\usepackage{enumitem}
\def\BibTeX{{\rm B\kern-.05em{\sc i\kern-.025em b}\kern-.08em
    T\kern-.1667em\lower.7ex\hbox{E}\kern-.125emX}}

\begin{document}

\title{Multi-Model Federated Learning}

\author{Neelkamal Bhuyan and Sharayu Moharir \\
Department of Electrical Engineering, Indian Institute of Technology Bombay\\
neelkamal.bhuyan@iitb.ac.in, sharayum@ee.iitb.ac.in
}

\maketitle

\begin{abstract}
Federated learning is a form of distributed learning with the key challenge being the non-identically distributed nature of the data in the participating clients. In this paper, we extend federated learning to the setting where multiple unrelated models are trained simultaneously. Specifically, every client is able to train any one of $M$ models at a time and the server maintains a model for each of the $M$ models which is typically a suitably averaged version of the model computed by the clients. We propose multiple policies for assigning learning tasks to clients over time. In the first policy, we extend the widely studied FedAvg to multi-model learning by allotting models to clients in an i.i.d. stochastic manner. In addition, we propose two new policies for client selection in a multi-model federated setting which make decisions based on current local losses for each client-model pair. We compare the performance of the policies on tasks involving synthetic and real-world data and characterize the performance of the proposed policies. The key take-away from our work is that the proposed multi-model policies perform better or at least as good as single model training using FedAvg.
\color{blue}
\end{abstract}

\begin{IEEEkeywords}
federated learning, multi-armed bandits, multi-objective optimization
\end{IEEEkeywords}

\section{Introduction}
Federated learning \cite{mcmahan2017communication} is a decentralized form of machine learning involving training a global model on local datasets of edge devices with the server only performing the function of aggregating client weight updates. The key feature in federated learning is that the local client datasets are never shared with each other or with the server and that these datasets are generally of non-i.i.d. nature.

Traditional machine learning, with its centralized training method, has faced issues with real-world deployment due to data privacy problems and its resource-intensive training process. Federated learning \cite{kairouz2019advances} is a solution to this. In fact, it has received a lot of attention in recent years in applied perspectives. For example, Google uses federated learning in their Gboard mobile keyboard \cite{hard2018federated, yang2018applied, chen2019federated, ramaswamy2019federated}. Apple is using cross-device FL in its mobile keyboard and voice assistant in iOS 13 \cite{appleFL}.

Several works have proposed client selection algorithms for learning a single task in federated setting \cite{8761315,ribero2020communication, chen2020optimal, xia2020multi, cho2020bandit}. However, to the best of our knowledge, the idea of learning multiple uncorrelated tasks simultaneously in a federated setting has not been explored yet. We present three client selection algorithms for training multiple uncorrelated models simultaneously in a federated learning setting. In addition to that, we also provide experimental results to show the behavior of simultaneous federated learning for multiple tasks.

We consider the setting where each client has datasets of $M$ uncorrelated models. These datasets are not necessarily i.i.d. in nature.
Each client is capable of training any of the $M$ models, however, due to \emph{computational constraints at the client}, it can train at most one model in each round. The server maintains $M$ aggregate models, one for each model.

We propose three algorithms - Multi-FedAvg, Ranklist-Multi-UCB, and Pareto-Multi-UCB for client selection and model assignment (for selected clients). Multi-FedAvg involves random selection of clients and random model assignment while the other two algorithms select clients by scoring them on their local loss. We report average test accuracy as the evaluation metric. Our experiments show that multiple models can indeed get trained in a federated setting. In fact, the performance of our multi-model algorithms is either better or at least as good as that of a single model FedAvg \cite{mcmahan2017communication} algorithm, under similar settings. This is highlighted by our experiments with synthetic as well as real-world tasks such as image classification of celebrity faces. In addition to that, our experiments show that under resource constraints (low number of clients per round), Ranklist-Multi-UCB and Pareto-Multi-UCB perform better than FedAvg with half the number of clients per round.

\section{Setting}

We consider the setting where the server trains $M$ unrelated models in a distributed manner using a pool for clients. Each client has a separate dataset for each model. The server maintains a global version of each of the $M$ models.
In each round, the algorithmic challenge at the the server is to allot at most one model to the clients for training.

At the start of each round, the server selects up to $K$ clients to be used for training. $K$ is a fixed parameter given as an input to our system. Each selected client receives global weights of the model it needs to train. The selected clients train the global version of the model allotted to them on the corresponding local training dataset and send the updated model weights to the server. The server weights are an average of the corresponding client weights with importance given to each client equal to the proportion of total training samples in it. The above method is the standard and most widely used form of training in federated learning. \cite{mcmahan2017communication, caldas2018leaf, cho2020bandit}. Refer to Algorithm \ref{alg4} for the pseudo-code of the training process and to Table \ref{tab1} for a description of variables used in the algorithms.

We evaluate the performance of the allocation policy using the value of the test accuracy of the models maintained by the server averaged over all agents as a function of time. To do this, the server communicates its models to the clients periodically and the clients test each model on the corresponding local test datasets. The details of the testing methodology are discussed in detail in subsequent sections.

\begin{algorithm}
\caption{Pseudo-code for $M$-model training at server}\label{alg4}
\hspace*{\algorithmicindent} \textbf{Input:} $\mathcal{S_T}$, $K$, policy $P$, $p_k(i)$ $\forall$ $k,i$\\
\hspace*{\algorithmicindent} \textbf{Initialize:} global model weights globalModel, local model weights localModel
\begin{algorithmic}[1]
\State globalModel$[m]$ $\gets$ \textbf{0} $\forall$ $m \in \{1,2,..,M\}$
\State Initialise parameters relevant to policy $P$ in round 0
\Repeat
\State localModel$[m,c]$ $\gets$ \textbf{0} $\forall$ $c \in \mathcal{S_T}, m \in \{1,2,..,M\}$
\State Update parameters relevant to policy $P$
\State ($\mathcal{S}^{(t)}$, trainModel) $\gets$ call function for $P$
\For{$c \in \mathcal{S}^{(t)}$}
    \State $m$ $\gets$ trainModel$[c]$
    \State localModel$[m,c]$ $\gets$ globalModel$[m]$
    \State Update localModel$[m,c]$ by local training
\EndFor
\For{$m \in \{1,2,..,M\}$}
    \State update $\gets$ $\frac{\displaystyle  \sum_{c \in \mathcal{S}^{(t)}}\text{localModel}[m,c]\times p_c(m)}{\displaystyle \sum_{c \in \mathcal{S}^{(t)}} p_c(m)\times \mathbbm{1}_{\{m=\text{trainModel}[c]\}}}$
    \State globalModel$[m]$ $\gets$ update
\EndFor
\Until{$n$ = max iterations}
\end{algorithmic}
\end{algorithm}

\section{Our Policies}
In this section, we discuss various policies for client selection and model assignment. 

\begin{table}
\caption{Common variables used in algorithms}
\begin{center}
\begin{tabular}{|c|c|}
\hline
\textbf{Variable Name} & \textbf{Description} \\
\hline
$\mathcal{S}_T$ & Set of all clients \\
$K$ & Number of clients per round\\
$\mathcal{A}_t$ & Set of all score vectors at round $t$, $\{A_t(k)\}_{k = 1}^N$\\
$\mathcal{S}^{(t)}$ & Set of selected clients for round $t$\\
trainModel & Models assigned to clients in $S^{(t)}$\\
$R$ & List of model ranklists\\
$n$ & round number\\
$M$ & Number of Models\\
$N$ & Total number of clients = $|\mathcal{S}_T|$\\
\hline
\end{tabular}
\label{tab1}
\end{center}
\end{table}

\subsection{Multi-FedAvg}
We first extend the popular FedAvg \cite{mcmahan2017communication} algorithm to training multiple models. FedAvg selects clients at random and averages the returned model weight updates by their dataset size. We keep these aspects of FedAvg. In addition to those, for each of the selected clients, the model to be trained is picked independently at random from one of the $M$ models. Refer to Algorithm \ref{alg1} for a formal definition of Multi-FedAvg.

\begin{algorithm}[h]
\caption{Pseudo-code for Mutli-FedAvg}\label{alg1}
\hspace*{\algorithmicindent} \textbf{Input:} $\mathcal{S_T}$, $K$\\
\hspace*{\algorithmicindent} \textbf{Initialize:}  $\mathcal{S}^{(t)} = \emptyset$, trainModel
\begin{algorithmic}[1]
\Procedure{Multi\_FedAvg}{$\mathcal{S_T}$, $K$}
\State select $K$ clients randomly from $\mathcal{S_T}$ to put in $\mathcal{S}^{(t)}$
\For{$c \in \mathcal{S}^{(t)}$}
    \State trainModel$[c] \gets$ randomly select from \{1,2,..., $M$\}
\EndFor
\State \textbf{return:} $\mathcal{S}^{(t)}$, trainModel$[c]$
\EndProcedure
\end{algorithmic}
\end{algorithm}
\subsection{Local Training Loss based Client Selection policies}
In this section, we introduce two new policies that use a scoring mechanism for selecting clients. Each client is assigned a score after every round which is used to select clients in the next round. We extend the scoring method used in \cite{cho2020bandit} which treats the client selection problem as a multi-arm bandit problem. It uses an UCB-index based on the discounted average local training loss, as the score for the clients.

We choose UCB index-based scoring as it pays importance to both reward maximization and exploration. The results presented in \cite{cho2020bandit} show that this form of biased client selection performs better than random selection and is fair to all clients. Below, we calculate this UCB-index for each model and have a $M$-length score vector for each client.

Here, $A_t(k)$ is the score vector for client $k$, $i$ denotes the model number and $t$ denotes the server round number. Local loss, for model $i$ of client $k$, is denoted by $l_t(k,i)$. The extent to which older training rounds influence the cumulative loss, $L_t(k,i)$, is controlled by $\gamma$ which lies between 0 and 1. Further, $p_k(i)$ is the proportion of model-$i$ training samples in client $k$ out of all model-$i$ training samples.
\begin{align}
A_{t}(k,i) &= p_k(i)(L_{t}(k,i)/N_{t}(k,i) + U_{t}(k,i))\label{eqn1} \\
\text{where, } L_{t}(k,i) &= \sum_{n = 0}^{t-1} \gamma^{t-1-n} \mathbbm{1}_{\{k\in \mathcal{S}^{(n)}\}} l_{n}(k,i)\label{eqn2}\\
N_{t}(k,i) &= \sum_{n = 0}^{t-1} \gamma^{t-1-n} \mathbbm{1}_{\{k\in \mathcal{S}^{(n)}\}} \label{eqn3}\\
\text{and }U_{t}(k,i) &= \sqrt{2\log \bigg(\sum_{n = 0}^{t-1} \gamma^{t-1-n}\bigg)/N_{t}(k,i)}\label{eqn4}. \\
\text{Further, }A_{t}(k) &= [A_{t}(k,1), A_{t}(k,2), ..., A_{t}(k,M)]^T. \label{eqn6} 
\end{align}

For $U_t$ to be defined, all clients' all models need to be trained at least once before policy-driven client selection.\\
Next, we discuss two client selection and model assignment policies based on the UCB-index score vector.

\subsubsection{Ranklist-Multi-UCB}
This policy selects clients in round $t>0$ as follows.
\begin{itemize}
    \item[--] Create a highest to lowest rank-list of score for each model, i.e., 
            $
                R(i) = sorted(\{A_t(k,i)\}_{k = 1}^N),
            $
 where $A_t(k,i)$ is as defined in \eqref{eqn1}.
    \item[--] Starting from the $((t \mod M) + 1)^{\text{th}}$ model, pick the topmost unpicked clients from the corresponding rank-lists in a round-robin manner and put them in the list of selected clients until the required number of clients per round are selected. If client number \textit{k} gets picked from rank-list \textit{i}, it is assigned model number \textit{i}. 
\end{itemize}
Refer to Algorithm \ref{alg2} for a formal definition. 

\begin{algorithm}
\caption{Pseudo-code for Ranklist-Multi-UCB}\label{alg2}
\hspace*{\algorithmicindent} \textbf{Input:} $\mathcal{A}_t, \mathcal{S_T}, K, n$\\
\hspace*{\algorithmicindent} \textbf{Initialize:} $\mathcal{S}^{(t)} = \emptyset$, trainModel, $R$
\begin{algorithmic}[1]
\Procedure{RM\_UCB}{$\mathcal{A}_t, \mathcal{S_T}, K, n$}
\For{$m \in \{1,2,..,M\}$}
    \State $R[m] \gets sorted(\{A_t(k,m)\}_{k = 1}^N)$\Comment{descending}
\EndFor
\State $start \gets n$
\State $count \gets 0$
\While{$|\mathcal{S}^{(t)}| \leq K$}
    \State $m \gets ((start + count) \bmod M) + 1$
    \State $c \gets$ Highest unpicked client in $R[m]$
    \State Put $c$ in $\mathcal{S}^{(t)}$ and trainModel$[c] \gets m$
    \State $count \gets count + 1$
\EndWhile
\State \textbf{return:} $\mathcal{S}^{(t)}$, trainModel$[c]$
\EndProcedure
\end{algorithmic}
\end{algorithm}

\subsubsection{Pareto-Multi-UCB}
\begin{enumerate}[label=\Alph*.]
    \item Client Selection: 
Recall that $\mathcal{A}_t$ denotes the list of score vectors at the beginning of round $t$. A client $c$ is a Pareto optimal client if $\mathcal{A}_t[c] \not\prec \mathcal{A}_t[j]$ $\forall$ $j \in \mathcal{S_T}$
    and $\mu \not\prec \nu$ if there is some dimension $i$ such that $\mu^i > \nu^i$. 
 We compute the set of all Pareto optimal clients \cite{drugan2013designing} of $\mathcal{A}_t$. 
    We call this set $\mathcal{A}_{pareto}$.
 We then obtain the set of clients whose score vectors belong to $\mathcal{A}_{pareto}$. Call it $\mathcal{S}_p$
 If number of clients ($= r$) is less than $|\mathcal{S}_p|$, sample $r$ clients from $\mathcal{S}_p$, else $\mathcal{S}_p$ is the set of clients for next round.
\item Model Assignment: We first create a highest to lowest rank-list of score for each model:
  $  R(i) = sorted(\{A_t(k,i)\}_{k = 1}^N). $
For each client, pick the model for which the client has the best rank in its associated rank-list.
\end{enumerate}
Refer to Algorithm \ref{alg3} for a formal definition. 

\begin{algorithm}
\caption{Pseudo-code for Pareto-Multi-UCB}\label{alg3}
\hspace*{\algorithmicindent} \textbf{Input:} $\mathcal{A}_t, \mathcal{S_T}, K$\\
\hspace*{\algorithmicindent} \textbf{Initialize:} $\mathcal{S}^{(t)} = \emptyset$, trainModel, $R$, empty set of pareto optimal clients $\mathcal{S}_p$
\begin{algorithmic}[1]
\Procedure{PM\_UCB}{$\mathcal{A}_t, \mathcal{S_T}, K$}
\For{$c \in \mathcal{S_T}$}
    \If{$\mathcal{A}_t[c]  \not\prec \mathcal{A}_t[j]$ $\forall j \in \mathcal{S_T}$}
        \State Put $c$ in $\mathcal{S}_p$
    \EndIf
\EndFor
\If{$|\mathcal{S}_p| \leq K$}
    \State $\mathcal{S}^{(t)} \gets \mathcal{S}_p$
\Else
    \State Pick $K$ clients randomly from $\mathcal{S}_p$ to put in $\mathcal{S}^{(t)}$
\EndIf
\For{$m \in \{1,2,..,M\}$}
    \State $R[m] \gets sorted(\{A_t(k,m)\}_{k = 1}^N)$\Comment{descending}
\EndFor
\For{$c \in \mathcal{S}^{(t)}$}
    \State trainModel$[c] \gets \displaystyle \argmin_{i \in \{1,2,..,M\}}$ rank of c in $R[i]$
\EndFor
\State \textbf{return:} $\mathcal{S}^{(t)}$, trainModel$[c]$
\EndProcedure
\end{algorithmic}
\end{algorithm}

\section{Experiment Details}
To ascertain the possibility of training multiple models simultaneously in a federated setting, we tested the three policies on tasks involving synthetic and real-world datasets. We were limited to training only two models simultaneously due to hardware constraints.
\subsection{Synthetic Datasets}
The policies were tested on Synthetic(1,1) \cite{li2018federated},\cite{li2019fair} and Synthetic-IID \cite{li2018federated},\cite{li2019fair} datasets with logistic regression as the classification model. Synthetic($\alpha$, $\beta$) \cite{li2018federated} is a dataset involving labelled feature vectors that can be non-identically or identically distributed across the clients. The variation in the underlying local models is decided by $\alpha$ while $\beta$ introduces non-i.i.d. nature in the local data of the devices. Synthetic-IID on the other hand has local data distributions identical for all devices. We consider two models, Model 1 is logistic regression classification of 60-dimensional vectors into 5 classes and Model 2 is logistic regression classification of 30-dimensional vectors into 10 classes.

\subsection{Real Dataset}
 We used the CelebA \cite{caldas2018leaf} dataset which is based on the Large-scale CelebFaces Attributes Dataset \cite{liu2015deep}. It has celebrity faces that have various binary attributes attached to them. The task associated with this dataset is a binary classification of the face images based on one of the attributes (fixed). In our experiment, we have Model 1 as binary classification into ``Smiling" vs ``Not Smiling" and Model 2 as binary classification into ``Eyeglasses present" vs ``Eyeglasses absent". Both tasks used Convolutional Neural Network (CNN) with one hidden layer.
\subsection{Framework}
We use a modified version of the LEAF Federated Learning Benchmarking framework \cite{caldas2018leaf} as the test-bench. LEAF is a single model FL framework. The additional features we added extend the framework to train two models simultaneously in a federated setting.

The server communicates with the clients for five hundred rounds while local training at a client is for one epoch. We evaluate the global models every ten communication rounds.

We report the average test accuracy as the metric to compare policies against each other. The average test accuracy is calculated in the following manner:
  The server version of both models is sent to all clients and is tested on the corresponding local test datasets.
  Each client's test accuracy is returned to the server.
  Each client is assigned a weight for each model. This weight is proportional to the total test data it hosts locally. The weighted average of the clients' test accuracy is reported as the average test accuracy.

We compare the performance of our policies to that of single model training using FedAvg. We consider this as the baseline for the evaluation of our proposed policies. Single model training implies that each model is trained on a separate run in the standard FL setting. The number of clients chosen per round for this is half of what is used for two-model training. This is assuming that in two model training, on average, the chosen clients in every round are equally distributed between the two models.

\section{Experiment Results}
\subsection{Synthetic Dataset}
\subsubsection{Synthetic(1,1)}
\begin{figure}
    \centering
  \subfloat[Test accuracy as a function of number of communication rounds for Model 1 \label{1a}]{%
   \includegraphics[width=0.75\linewidth]{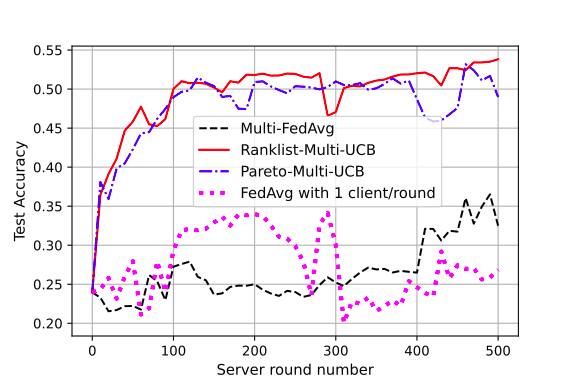}}
    \hspace{2pt}
    \subfloat[Test accuracy as a function of number of communication rounds for Model 2 \label{1b}]{%
   \includegraphics[width=0.75\linewidth]{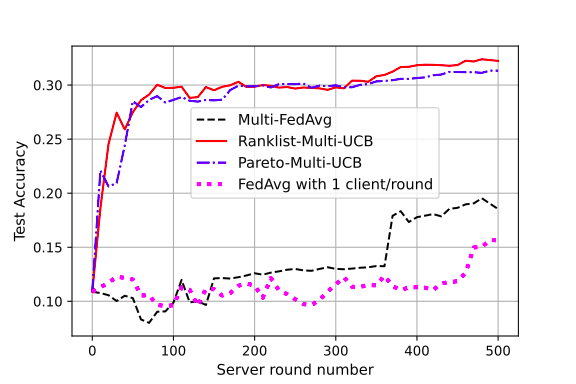}}
  \caption{Synthetic(1,1) with 2 clients per round. Ranklist-Multi-UCB and Pareto-Multi-UCB perform significantly better than FedAvg with 1 client/round, in both models}
  \label{fig1} 
\end{figure}

\begin{figure}
    \centering
  \subfloat[Test accuracy as a function of number of communication rounds for Model 1 \label{3a}]{%
   \includegraphics[width=0.75\linewidth]{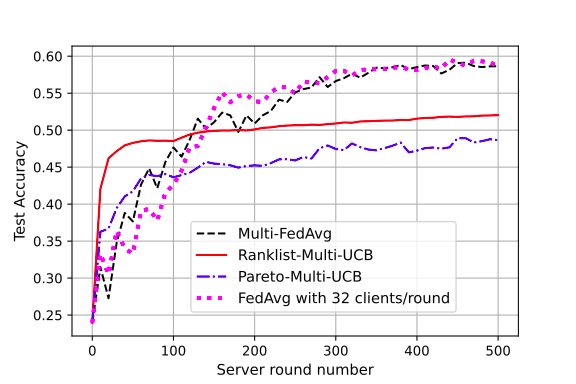}}
    \hspace{2pt}
    \subfloat[Test accuracy as a function of number of communication rounds for Model 2 \label{3b}]{%
   \includegraphics[width=0.75\linewidth]{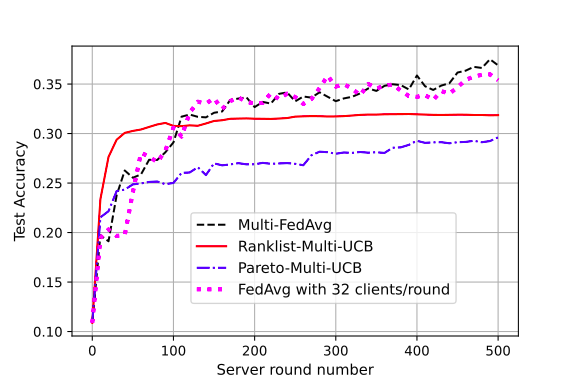}}
  \caption{Synthetic(1,1) with 64 clients per round. Multi-FedAvg performs as good as FedAvg with 32 clients/round}
  \label{fig3} 
\end{figure}

\begin{figure}
    \centering
  \subfloat[Test accuracy as function of number of clients per round for Model 1 \label{6a}]{
   \includegraphics[width=0.75\linewidth]{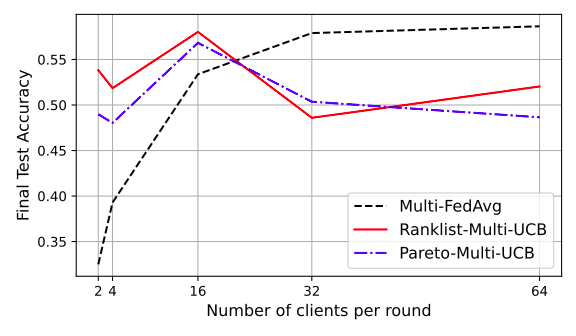}}
    \hspace{2pt}
    \subfloat[Test accuracy as function of number of clients per round for Model 2 \label{6b}]{
  \includegraphics[width=0.75\linewidth]{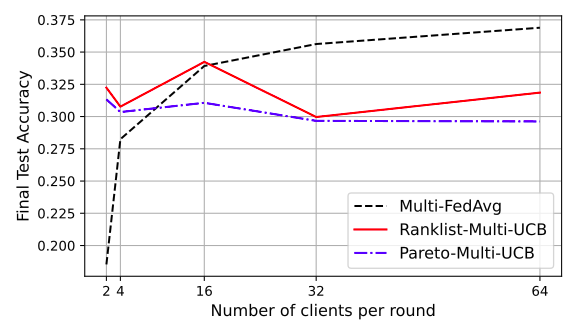}}
  \caption{Variation of test accuracy with number of clients per round in Synthetic(1,1). When the number of clients per round is low, score-based policies perform better whereas for higher clients per round Multi-FedAvg performs better.}
    \label{fig6}
\end{figure}

In the case of two clients per round, Fig. \ref{fig1} shows that Ranklist-Multi-UCB and Pareto-Multi-UCB perform better than FedAvg with one client per round. Multi-FedAvg seems to be at the same level as FedAvg with one client per round. It is also observed that Ranklist-Multi-UCB and Pareto-Multi-UCB see an instant leap in test accuracy in the initial rounds whereas Multi-FedAvg and FedAvg show slow growth in this phase.\\
In the case of sixty-four clients per round, we observe in Fig. \ref{fig3} that Multi-FedAvg performs better than the two local loss-based policies. In addition to that Multi-FeadAvg performs as well as FedAvg with half the clients, for both models. During the initial rounds, Ranklist-Multi-UCB and Pareto-Multi-UCB have higher test accuracy whereas, in the long run, Multi-FedAvg achieves higher test accuracy.
Variation of test accuracy with the number of clients per round in Fig. \ref{fig6} shows that in Synthetic(1,1) our score-based policies perform better than Multi-FedAvg in case of inadequate clients per round but worse when clients per rounds are sufficient.


\subsubsection{Synthetic-IID}
\begin{figure}
    \centering
  \subfloat[Test accuracy as a function of number of communication rounds for Model 1 \label{2a}]{%
      \includegraphics[width=0.75\linewidth]{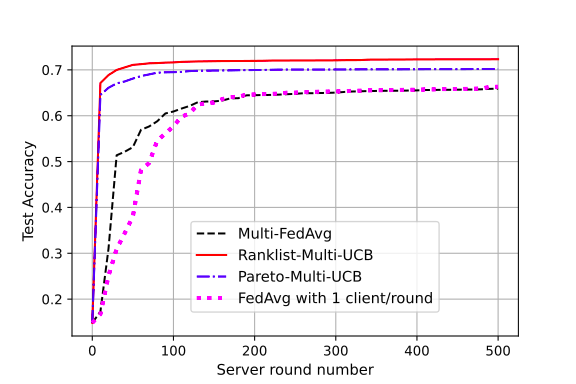}}
    \hspace{2pt}
    \subfloat[Test accuracy as a function of number of communication rounds for Model 2 \label{2b}]{%
  \includegraphics[width=0.75\linewidth]{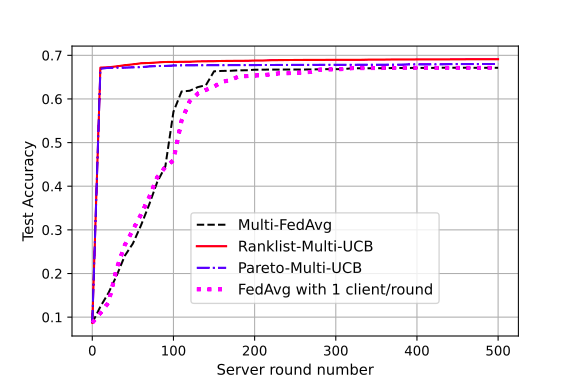}}
  \caption{Synthetic-IID with 2 clients per round. Ranklist-Multi-UCB and Pareto-Multi-UCB perform better than FedAvg with 1 client/round, in both models}
  \label{fig2} 
\end{figure}

\begin{figure}
    \centering
  \subfloat[Test accuracy as a function of number of communication rounds for Model 1 \label{4a}]{%
      \includegraphics[width=0.75\linewidth]{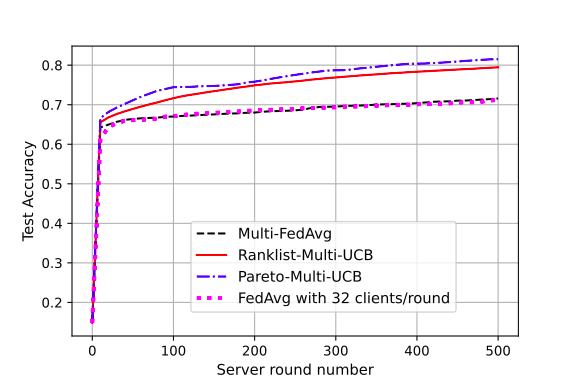}}
    \hspace{2pt}
    \subfloat[Test accuracy as a function of number of communication rounds for Model 2 \label{4b}]{%
  \includegraphics[width=0.75\linewidth]{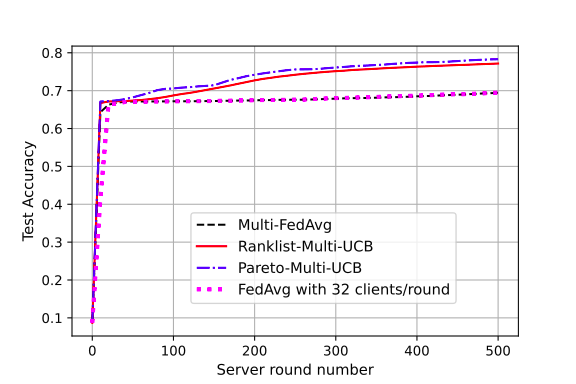}}
  \caption{Synthetic-IID with 64 clients per round. Ranklist-Multi-UCB and Pareto-Multi-UCB perform better than FedAvg with 32 clients/round, in both models}
  \label{fig4} 
\end{figure}

\begin{figure}
    \centering
  \subfloat[Test accuracy as function of number of clients per round for Model 1 \label{7a}]{%
  \includegraphics[width=0.75\linewidth]{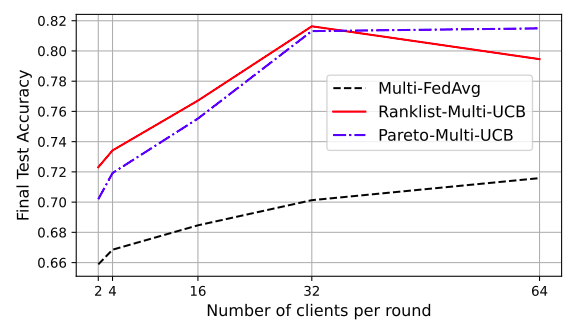}}
    \hspace{2pt}
    \subfloat[Test accuracy as function of number of clients per round for Model 2 \label{7b}]{%
  \includegraphics[width=0.75\linewidth]{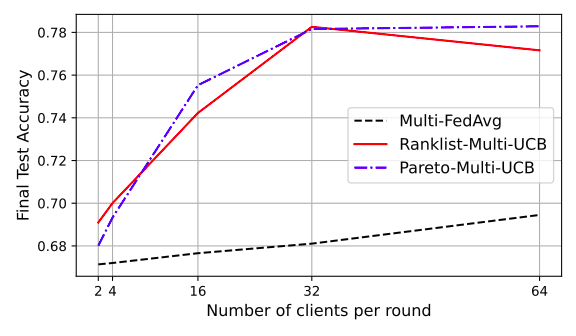}}
  \caption{Variation of test accuracy with number of clients per round in Synthetic-IID. For low or high number of clients per round, score-based policies perform better than Multi-FedAvg}
    \label{fig7}
\end{figure}

In the case of two clients per round, Fig. \ref{fig2} shows Ranklist-Multi-UCB and Pareto-Multi-UCB again outperform Multi-FedAvg which has performance similar to FedAvg with one client per round. However, the gap in test accuracy between Multi-FedAvg and the other two is less compared to the non-i.i.d. datasets.\\
In the case of sixty four clients per round, Fig. \ref{fig4} shows that the performance of the policies follows a similar trend to their counterparts when the number of clients per round is 2.  Ranklist-Multi-UCB and Pareto-Multi-UCB perform better than Multi-FedAvg and FedAvg in both models. The gap in test accuracy in Model 2 for the three policies is more than when the number of clients per round is 2.\\
Variation of test accuracy with number of clients per round in Fig. \ref{fig7} shows that in Synthetic-IID, our policies perform better than Multi-FedAvg under all scenarios.

\subsection{CelebA Dataset}
\subsubsection{Small number of clients (2) per round}
\begin{figure}
    \centering
 \subfloat[Test accuracy as a function of number of communication rounds for Model 1 \label{8a}]{%
      \includegraphics[width=0.75\linewidth]{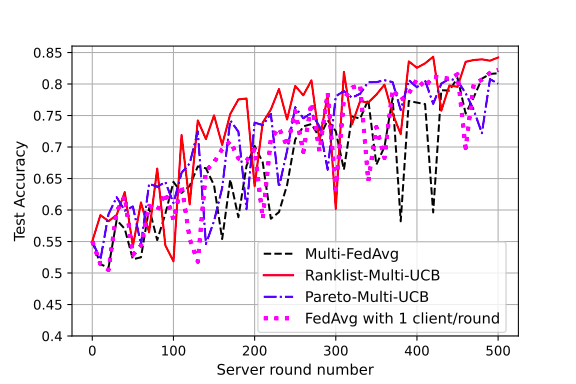}}
    \hspace{2pt}
    \subfloat[Test accuracy as a function of number of communication rounds for Model 2 \label{8b}]{%
  \includegraphics[width=0.75\linewidth]{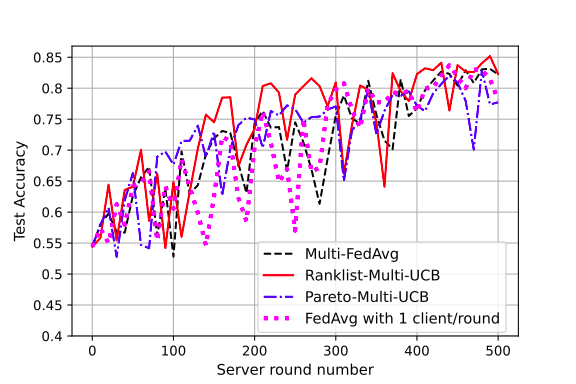}}
 \caption{CelebA with 2 clients per round. Ranklist-Multi-UCB seems to be performing marginally better than the other policies}
 \label{fig8} 
\end{figure}

For two clients per round, the test accuracy trend for all three policies is noisy, as seen in Fig. \ref{fig8}. However, Ranklist-Multi-UCB seems marginally better than Multi-FedAvg, Pareto-Multi-UCB, and FedAvg with one client per round. It is observed that test accuracy in the case of Ranklist-Multi-UCB is, in general, slightly higher than that of the other two policies, especially for Model 1.
\subsubsection{Large number of clients (10) per round}
\begin{figure}
    \centering
  \subfloat[Test accuracy as a function of number of communication rounds for Model 1 \label{5a}]{%
   \includegraphics[width=0.75\linewidth]{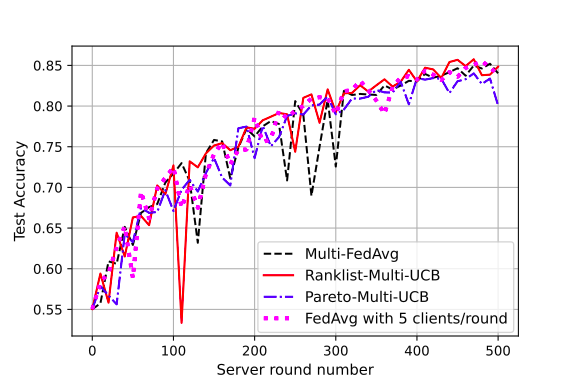}}
    \hspace{2pt}
    \subfloat[Test accuracy as a function of number of communication rounds for Model 2 \label{5b}]{%
   \includegraphics[width=0.75\linewidth]{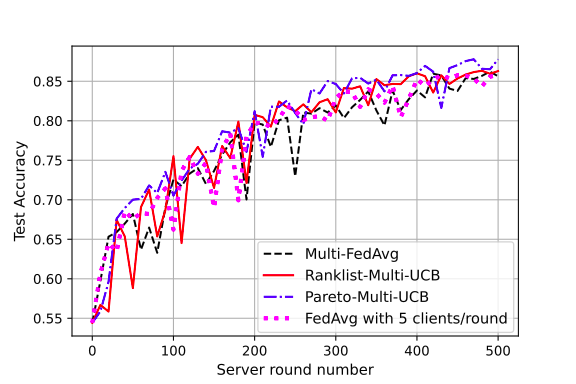}}
  \caption{CelebA with 10 clients per round. All four policies have similar performance in both models}
  \label{fig5} 
\end{figure}

For ten clients per round, Fig. \ref{fig5} shows that the overall performance of all three policies is similar to that of FedAvg with five clients per round. However, we observe that Multi-FedAvg and Ranklist-Multi-UCB have a more noisy test accuracy trend compared to Pareto-Multi-UCB. Variation of test accuracy with number of clients per round 
do not reveal any significant trend other than all policies performing similarly.

\section{Conclusions and Future Work}
In this study, we demonstrate that multiple unrelated models can be trained simultaneously in a federated setting. We extend the established FedAvg algorithm to multi-model training. In addition to that, we propose two new policies called Ranklist-Multi-UCB and Pareto-Multi-UCB for client selection in multi-model training in the federated setting. We compare the performance of these three policies with single model training using FedAvg. This is done on both synthetic and real-world datasets. We find that when the number of clients per round is low, Ranklist-Multi-UCB and Pareto-Multi-UCB outperform single model FedAvg. In addition to that, we find that performance of Multi-FedAvg is similar to FedAvg under all our test scenarios. It is important to note that our multi-model policies perform at least as well as single model FedAvg. Interestingly, and when the number of clients per round is low, our multi-model policies even outperform single model FedAvg. The tasks involved in CelebA dataset were image classification, which is a common application of machine learning on edge devices. We successfully showed that multi-model federated learning is viable in real-world tasks with results at least as good as when these models are trained independently.

Potnetial future work directions include incorporating features like client unavailability and constraints on clients' training capabilities to explore multi-model federated learning in more complex scenarios.

\bibliography{refs}
\bibliographystyle{IEEEtran}

\end{document}